\documentclass[10pt,letterpaper]{article}

\usepackage{cogsci}

\cogscifinalcopy 

\usepackage[
    backend=biber,
    style=apa,
    natbib=true,
    doi=false,
    isbn=false,
    url=false,
]{biblatex}
\usepackage{float} 

\usepackage{amsmath}
\usepackage{graphicx}
\usepackage{color}
\usepackage{lipsum}
\usepackage{todonotes}

\title{GPT-ology, Computational Models, Silicon Sampling: \\How should we think about LLMs in Cognitive Science?}
 
\author{{\large \bf Desmond C. Ong} (desmond.ong@utexas.edu)
\\
  Department of Psychology, The University of Texas at Austin
  }

\begin{document}

\maketitle

\begin{abstract}
 
Large Language Models have taken the cognitive science world by storm. It is perhaps timely now to take stock of the various research paradigms that have been used to make scientific inferences about ``cognition" in these models or about human cognition. We review several emerging research paradigms---GPT-ology, LLMs-as-computational-models, and ``silicon sampling"---
and review recent papers that have used LLMs under these paradigms. In doing so, we discuss their claims as well as challenges to scientific inference under these various paradigms. We highlight several outstanding issues about LLMs that have to be addressed to push our science forward: closed-source vs open-sourced models; (the lack of visibility of) training data; and reproducibility in LLM research, including forming conventions on new task ``hyperparameters" like instructions and prompts.

\end{abstract}


\section{Introduction}
Recent scientific discourse in cognitive science in 2023 and 2024 seems to be all about Large Language Models (LLMs) \citep{binz2023using, trott2023large}, including entire workshops at CogSci2023\footnote{https://cogscillm.com/} \citep{hardy2023large} and similar conferences. Over this short time, we have seen a variety of different research paradigms emerge. Some research is focused on evaluating the cognitive capacities of these LLM models: what they can or cannot do. One might call this type of research ``\textbf{GPT-ology}", which involves studying how LLMs like GPT-4 
process information\footnote{This name is inspired by earlier work on studying how BERT, then the most successful NLP model, worked, which earned the moniker ``BERT-ology" \citep{rogers2021primer}.}. These papers are usually characterized by inferences about the state of artificial cognition, and less on insights for human cognition. 
Other researchers suggest that we should instead abstract away from specific models, and use \textbf{LLMs as a computational model} of human cognition (e.g., \cite{blank2023large, frank2023openly}). 
One example within this paradigm uses LLM performance as an ``Existence Proof" about the sufficient conditions for certain cognitive capabilities to emerge \citep{contreras2023large, kauf2023event, piantadosi2023modern}. 
And finally, some researchers have been using LLMs to simulate human behavior \citep{argyle2023out, park2023generative, dillion2023can, grossmann2023ai}, and exploring how we can use these ``\textbf{silicon samples}" to make inferences about people.


\begin{figure*}[tb]
    \begin{center}
    \includegraphics[width=1\textwidth]{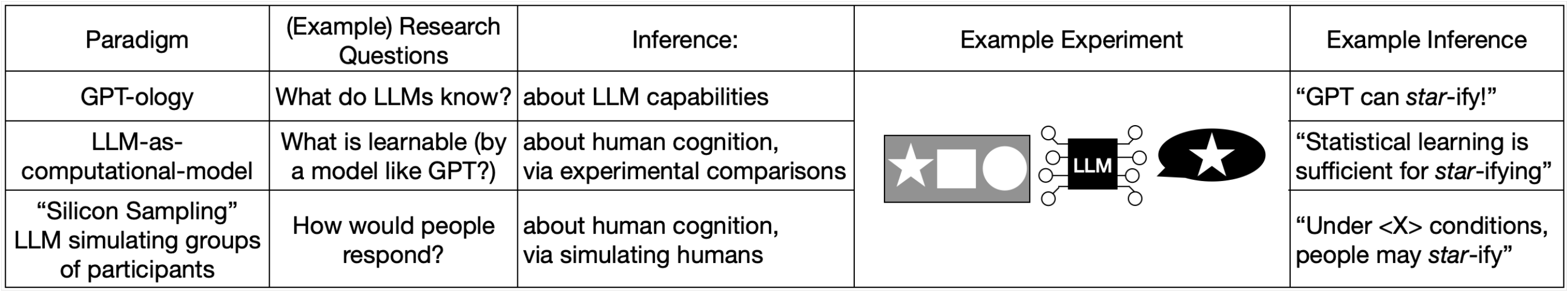}
    \end{center}
    \caption{
    The same experiment, assessing LLM performance on a given task---in this cartoon, presenting an LLM with a choice, and the LLM output is \emph{star}---leads to different inferences based on the initial research questions. 
    Researchers may make inferences about the capabilities of specific LLMs (``GPT-ology"), such as: ``GPT can \emph{star}-ify". Alternatively, we could use LLMs as a computational model of human learning. One example inference that could be made is that ``statistical learning alone is sufficient for \emph{star}-ifying". And finally, we could treat samples from an LLM under some conditioning contexts as illustrative of how people might respond in that manner (``under $<$\emph{X}$>$ conditions, people may \emph{star}-ify"). We note that these paradigms are not exhaustive (more creative ones could appear), nor are they mutually exclusive; the same paper or research program could make various claims.} 
    \label{paradigm-figure}
\end{figure*}


Given the interest in LLMs in cognitive science, we thought it timely to take stock of progress in these research paradigms. These paradigms are not exhaustive; In fact, we fully expect more creative approaches to appear. Neither are they mutually exclusive: the same empirical evidence---having an LLM respond to questions like in a psychological experiment---could be used to support research questions in more than one paradigm (see Fig. \ref{paradigm-figure}). 
The goal of this paper is to lay out a framework for thinking about these different approaches to using or studying LLMs. This is important because, in laying out these research questions at a ``bird's-eye" level, we can discuss the outstanding issues that affect most, if not all, of these research efforts, and that we feel have to be addressed as a field. We do not claim to have answers to all of these issues, although we provide some thoughts and suggestions for resolutions. 


In this short review, we focus on examining the scientific logic and assumptions inherent in these approaches.  
%
%
Due to the ``meta"-level of this discussion, we will not comment on specific psychological claims---for instance, claims about specific linguistic capabilities---vis-\`a-vis the literature and background knowledge required for adjudicating those claims. Instead, we will focus our discussion on more general principles: scientific reliability and validity, logical deductions, and epistemic support.

\section{A typology of research paradigms}

Here, we define a typology of research paradigms that have been applied to study LLMs. These paradigms are not mutually exclusive; in fact, they all rely on the same base experiment of having LLMs provide responses to some input (e.g., see Fig. \ref{paradigm-figure}), and so the same data can be used to address multiple paradigms. We differentiate these paradigms by their specific research goals, and hence the inferences that are made from the results. 


For each paradigm, we lay out some research questions, typical experiments, results and inferences that are made, and importantly, some of the challenges and concerns about interpretation that should be addressed.

\subsection{GPT-ology: Making inferences about LLMs}

There have been much effort focused on evaluating the cognitive capacities of LLMs, in order to draw inferences 
about the capabilities of certain models, either in isolation or relative to other LLMs \citep{hagendorff2023human}.



\textbf{LLM ``traits".} Some researchers have LLMs respond to validated psychological scales, and make inferences about their ``personality" or ``psychometrics" \citep{pellert2024ai, separiogarcia2023personality}. For instance, \citet{schaaff2023exploring} had chatGPT fill in empathy and autism-related scales, and concluded that ``the empathic abilities of chatGPT are still below the average of healthy [neurotypical] humans".

\textbf{Bias in LLMs.} Researchers have also studied potential biases that may be present in LLMs, and that may carry over to other contexts, such as when LLMs are generating answers.
Various groups \citep{tao2023auditing, abdurahman2023perils, fischer2023does} had GPT respond to surveys (e.g., the World Values Survey), and compared the responses to different cultural groups and other subgroups (e.g., political orientation), as a way to identify cultural or other biases in LLMs (e.g., ``\emph{GPT exhibits cultural values resembling [more Western,] Protestant countries}"; \citealp{tao2023auditing}). In a similar vein, \citet{hu2023generative} had LLMs complete sentences and showed that these models exhibit social identity biases. 

\textbf{LLMs as lab participants.} 
In addition to survey questions, LLMs can respond to tasks that one might provide in a typical cognitive science experiment. For instance, \citet{binz2023using} assessed GPT-3's performance on a variety of standard cognitive psychology tasks. 
Similar experiments have been done for tasks like analogical reasoning \citep{webb2023emergent}, logical reasoning \citep{lampinen2023language}, and inductive reasoning \citep{han2024inductive}.
A specific task that has attracted much interest is whether LLMs have ``Theory of Mind", operationalized as whether they can correctly answer questions that require representing others' false beliefs and desires \citep{kosinski2023theory, sap2022neural, gandhi2023understanding, trott2023large, ullman2023large}.

\textbf{Other LLM capabilities.} There are also many examples beyond standard cognitive tasks. For instance, several groups have looked at whether LLMs can generate ``empathic" responses \citep{yin2024ai, lee2024large, li2024skill, zhan2024large}.

\subsubsection{Challenges.} One obvious challenge, especially with regard to inferences about ``traits", is anthropomorphism. Many psychologists would intuitively reject the notion of an AI having ``personality" or other traits in the same way a human does\footnote{Although there is a longstanding and sizable research community in AI working on creating social virtual agents with distinct or tuneable ``personalities"}, and it is unclear even if these properties reliably affect behavior (e.g., generated output text) in any systematic fashion. 

Another challenge is reliability (see also Outstanding Issues 3 and 5 below). LLMs are not static in the sense that their knowledge depends upon their training data, and commercial models like GPT are updated regularly, and without the ability to use past checkpoints. Re-testing the model after an update (or even the next major ``version", like GPT5) might already change the results, making them obsolete. 





But reliability may be difficult to achieve even in the short-term.  
LLM performance is susceptible to seemingly innocuous changes in the stimuli prompts, such as reversing the order of answers (in a multiple choice prompt) or information \citep{binz2023using}, or trivial alterations to the stimuli \citep{ullman2023large}. It is not quite clear why LLM performance is so brittle to the specific prompt. One possible (and scientifically uninteresting) answer is that the specific task (or tasks like it) were present in the training data, and the LLM has simply memorized its answer, and thus is unable to handle superficial changes. This will have to be answered for proper interpretation.


\subsection{LLMs as computational models}

For the previous research paradigm(s), the inference is about the LLM itself, such as whether a particular LLM possesses some reasoning capability or bias. Another set of research questions aims to make inferences about human cognition, by assuming that, at some level, the LLM may be a computational model of human cognition. This is not an unfamiliar paradigm; cognitive scientists have long been making inferences about human learning and cognition via comparison with much simpler models (e.g., feed-forward neural networks; rule-based systems, agent simulations). The main difference with current LLMs is the sheer scale and capabilities of LLMs compared to previous generations of models, which opens up new possibilities for experimentation, and consequently, new inferences about learning and reasoning. There is still no consensus about how to appropriately compare LLMs to human cognition, although many researchers are writing about it \citep{blank2023large, frank2023openly}. For example, borrowing Marr's levels of analysis, are we comparing LLMs and human performance at the computational level \citep{blank2023large}? Or can we make inferences about processing streams in LLMs versus humans, even at a neural level \citep{hosseini2024artificial}?

One approach is to compare the performance and pattern of errors of LLMs with human performance (including ``biases" and errors), which might yield interesting insights into how these errors might be learnt from text data \citep{aher2023using}. For instance, GPT-3 makes Kahneman and Tversky (\citeyear{kahneman1972subjective})-esque errors, such as the conjunction fallacy, rating the probability that ``Linda is a feminist bank teller" larger than ``Linda is a bank teller" \citep{binz2023using}.
Other work has also looked at whether LLMs make similar moral acceptability judgments as humans \citep{dillion2023can, jin2022make}. At the moment it is difficult to understand ``why" these LLMs make these patterns of human-like decisions and errors, although perhaps there will be progress made here, such as by, for example, using techniques from developmental psychology \citep{frank2023baby}.


\textbf{Existence Proofs.} For some research questions, the mere demonstration of an LLM's capabilities serves as an Existence Proof about the learnability of some aspects cognition. For instance, how much of language can be learnt versus has to be innate \citep{contreras2023large, piantadosi2023modern}, or the gap between language and thought \citep{mahowald2024dissociating}. 
The general idea is that because we know, at a coarse level, what LLMs are exposed to during training, this might allow answering questions around the sufficient conditions for learning.
For instance, even though LLMs are only trained on language data, could they learn event knowledge (about whether one event is more likely than another), solely based on the statistics of word co-occurrences in its training data
\citep{kauf2023event}?
Or how does nonsymbolic learning from natural language give rise to symbolic reasoning
\citep{geiger2023relational}? 
In a recent example that includes language and visual input, \citet{vong2024grounded} trained a neural network with unlabeled audio and visual input from a single infant-worn head-camera taken over several months, and found evidence for grounded language acquisition from statistical learning alone.

\textbf{Internal representations.} And finally, we could theoretically peer into the inner workings of these models to see how they `think'. There are lots of ``probing" and other introspection methods developed in NLP to study the internal representations of such models as they learn \citep{belinkov2022probing}.
One could also peer into individual `neuronal' activation, or patterns of activation, and perhaps compare that with human brain activity (\cite{kumar2022reconstructing, hosseini2024artificial}, see also \cite{yamins2014performance} for the visual cortex and Computer Vision models), to make inferences. One could also run causal interventions or other mechanistic interpretability analyses on the model itself to test how information is processed in the model \citep{yamakoshi2023causal, wu2024interpretability}.

\subsubsection{Challenges.} When comparing LLM outputs to human behavior, there are many issues to consider. First, how much of the behaviors (e.g., errors like the conjunction fallacy) are due to them being present in the training data \citep{binz2023using}, or due to specific styles of prompting (see Outstanding Issues 4 and 5 below)?

The existence proof logic is asymmetric, especially with respect to failures (null results): if a model can ``do X", we could make a claim about the sufficient (but not necessary) conditions for a capability to emerge. But if an LLM model ``cannot do X", that cannot be used as an argument for the necessary conditions (that the LLM lacks) for that capability. For instance, 
LLM failure on Winograd Schema tasks might lead to an inference that world knowledge or commonsense knowledge is necessary, or that statistical learning from language co-occurences by itself is insufficient. But these inferences do not logically follow, and could easily be falsified with a newer and more capable model. 
Indeed, AI development over the past decade seems to be accelerating faster than many expect. 
Many researchers have catalogued the current inadequacies of LLMs---for example, failing to do certain types of reasoning or logic (e.g., \citealp{borji2023categorical})---and it is important to do so. 
But these null results do not yet lend themselves well to lasting scientific inferences, if all it takes is an engineering counterproof. Arguing from a lack of ability is less scientifically sound than an argument from the presence of one. (See also Outstanding Issue 2, below)

Studying internal representations of LLMs may not be possible, especially with proprietary models like the GPT-series models, which are only accessible via a limited API (see Outstanding Issue 3).









\subsection{Silicon Sampling: LLMs simulating humans}


Another approach that has gotten some attention is using LLMs to simulate populations or subgroups of humans \citep{grossmann2023ai, dillion2023can}, which has sometimes been referred to as ``silicon sampling" \citep{argyle2023out}. Researchers have used LLMs to simulate human behavior, for example in economic experiments \citep{aher2023using, horton2023large} or consumer preferences \citep{sarstedt2024using}. Researchers have also looked at specific subgroups, by conditioning the model with backstories of different subpopulations, and showed that LLMs could predict behaviors like voting \citep{argyle2023out}. To the extent that LLMs accurately ``encode" or ``compress" human knowledge, and that conditioning the model to reproduce the behavior of certain types of humans yields behavior of sufficient fidelity (two very strong assumptions), this approach might provide a scalable way to study human or ``human-like" cognition and behavior \citep{park2023generative}.

\subsubsection{Challenges.} This approach rests on several assumptions of fidelity, which is broadly whether LLM responses can accurately reflect human responses \citep{argyle2023out, grossmann2023ai}. These have to be properly tested, and are also related to issues with LLM reliability (See Outstanding Issues 3, 5 below). 

One appeal of this approach is that we could use LLMs to simulate subpopulations that might be more difficult to recruit in traditional studies, such as minority groups. But these minority groups are also underrepresented in the training data (See Outstanding Issue 4). 
Relatedly, there are concerns about whether such simulated behavior might reflect (biased) \textit{stereotypes} about how certain groups of people behave, rather than actual behavior. 

And lastly, this approach assumes that LLMs can simulate individual humans---or rather, LLM outputs are samples from an underlying distribution that might be in some way a good approximation to real human distributions. But other approaches view LLMs output as more of a population average, or ``aggregate" summary of human knowledge
(e.g., a cultural technology like a library; \citealp{yiu2023transmission}). These two conceptualizations are distinct and will affect experimental design and inferences drawn---an analogy in Bayesian cognitive science is whether people are sampling from a posterior distribution (probability matching), or whether people are reasoning using the \emph{maximum a posteriori} estimate. If LLMs are actually representing some kind of population average, but are treated as mimicking individual humans, this might lead to biases in the inferences drawn from these results.


\subsection{Other uses of LLMs}

We end this section with a brief mention of several other approaches that could become more developed in the future. First, LLMs can serve as building blocks in more complex models of cognition. For instance, LLMs can be used to extract features from unstructured text, as part of a larger neurosymbolic model \citep{kwon2023neuro, zhan2023evaluating}. Second, LLMs also have broad applicability in other aspects of psychological research \citep{demszky2023using}, such as to generate, classify, or annotate stimuli \citep{rathje2023gpt, ziems2023can}. Many of the challenges and outstanding issues (about reliability or validity) may also apply to these use-cases. 


\section{Outstanding issues}





\subsection{1. Which LLMs should we use?}

There exists a veritable zoo of language models (e.g., LLaMA, Alpaca, Vicuna), and most variants also come with different ``version numbers" (GPT-3, chatGPT, GPT-4) and sizes (LLaMA-7B, -13B, etc.). Some are open-sourced, while others are proprietary. Which should we be using?
For cognitive science research, should we just focus on one model, perhaps, the biggest---or more realistically, the best that one can have access to and can afford? This may introduce another layer of inequality as well-resourced labs may have greater access to unreleased or more expensive models. Should we instead be focused on experimenting with a range of models, as is done in machine learning research? 
How do these choices affect reproducibility?



\subsection{2. What inferences should we make if one LLM, but not others, can “do X”?}


Scientifically, the breadth of LLM choice poses an interesting conundrum. What should we make of contexts when ``smaller models" fail at a certain task, while ``bigger/better models" succeed \citep{gandhi2023understanding, kosinski2023theory, hagendorff2023human}, For instance, if a particular ``ability" was demonstrated by GPT-4 but not GPT-3. Is there some cognitively interesting answer about the model, learning algorithm, or data that leads to those changes? How does that affect arguments about the sufficiency of statistical learning or other conditions? Unfortunately, those questions seem like they would yield engineering answers, rather than cognitive insights. Relatedly, what if a particular cognitive ``ability" was restricted to one particular model (say, LLaMa-3), but not shared by other models of similar specifications? We think there are deep meta-scientific conversations that we could have as a field, rather than only in peer-review.


\subsection{3. LLMs are proprietary commercial products updated by companies.}

This issue contributes to many challenges already described earlier. Many of the papers we reviewed used closed-source proprietary models, notably GPT-3 or GPT-4. Closed-source means we do not have access to the model and data that can help guide inferences \citep{frank2023openly}, via introspecting model activations or understanding trends in training data. Moreover, the fact that companies regularly update their models (and perhaps even learn from previous input) might render the idea of reproducibility meaningless. Researchers might lose access to these models for a variety of sudden, unforeseeable reasons, such as economic, political, or legal (there are pending lawsuits and legislation in several jurisdictions). 

This brings up a deeper question that we should be asking as a field: Should we really be yoking the success of our science to such commercial products? Of course, commercial products are important to science, providing services (e.g., Qualtrics and other software, compute) and equipment necessary for scientific research. But these conditions are different, where the actual research artefact, the object of study, is a commercial product that is not regulated and that researchers have little influence over. 

Another concern about propriety models is the lack of transparency around engineering changes that are built into the model. To minimize liability concerns, many commercial LLMs have what are called ``\textbf{guardrails}" built into their system. For instance, GPT will refuse to discuss dangerous (e.g., making weapons), illegal (e.g., abuse), or offensive (e.g., racist jokes) information. But some of these guardrails might also affect research, for example, constraining a model's moral judgments to conform to a particular view. Some of these guardrails (or perhaps due to human feedback in training) may also result in idiosyncratic behaviors: for example, when asked to generate first-person negative stories, GPT tends to offer a happy ending (or a ``moral of the story"). Without more transparency, it is unclear which of these behaviors are learnt from data and which are engineered into the model.

One solution is to move to open-source models, but we as a field would still have to standardize many conventions (e.g., which model, Issue 1, or reproducibility, Issue 5).

\subsection{4. Training Data}
Training data is important for making claims about learnability, and also for ``simulating humans". However, the large amount of training data presents serious issues for scientific inference. First, for some proprietary models, the source and types of training data are not public information (e.g., GPT-4). Second, even if they were public, the sheer scale of data makes it difficult to assess (let alone control) what went into a model. Third, the data the model is trained on might contain biases that subsequently will affect its output.

One particular concern is \textbf{data leakage}. If an experimental task happens to be in the training data (e.g., the ``Sally-Anne" False-Belief task, or Kahneman-and-Tversky-style fallacy items), then the model might succeed on the task simply from having seen and memorized it in its training data. 
Memorization is a much less interesting scientific explanation for cognitive performance on a task, but it is often a concern, given that for example, trivial alterations to the stimuli like word order or changing names can break LLM performance on a task \citep{binz2023using, ullman2023large}. If we cannot guarantee that our tasks were not part of the training data, then a large portion of our experimental approach will be rendered invalid.

It is also worrying that LLM-produced output may form the training data for future generations of LLMs. Is human-written language on the internet like the Ship of Theseus, gradually being replaced by LLM-written approximations of human text? At what point might such language be no longer ``human"?

Lastly, LLMs are trained on data that predominantly comes from Western, Educated, Industrialized, Rich, and Democratic (WEIRD; \citealp{henrich2010weirdest, henrich2023cultural}) countries, and even within WEIRD societies, specific subcultures (mostly young, internet-savvy users) \citep{abdurahman2023perils, tao2023auditing, henrich2023cultural}. This is not representative and is worrying if LLM-based cognitive science becomes mainstream.

\subsection{5. Reproducibility in LLM research}

Issue 3 makes it impossible to guarantee continued access to a stable, dated version of a proprietary model. But even for open-sourced models, there are other issues.

\textbf{Stochasticity} is a feature of language (modeling). Instead of always returning the same sequence of words, LLMs sample words probabilistically from a distribution, and the randomness is controlled by a parameter called the temperature. 
A common misconception is that the stochasticity of LLMs is a flaw for reproducibility, and some studies are run with temperature set to 0 \citep{binz2023using}. 
This is an incorrect perception: stochasticity is a \emph{feature}, not a bug, and is true also of human cognition more generally---humans do not always give the same answer either, but psychologists have learnt to sample from people. 
Ideally we would be measuring (and making inferences over) the \textit{probability distribution of tokens} in LLMs. This is directly observable with some open-source LLMs, but not always true with proprietary LLMs. 
This property suggests that, in many experimental tasks, scientists should be thinking about \textbf{collecting samples from LLMs} (just like how we sample responses from many people, or even the same person multiple times), and \textbf{doing statistics over those samples}. This is currently not common practice.

\textbf{Prompting} or prompt engineering is the process of iterating and deciding the best natural language input to an LLM to increase the performance of the output. This is more an ``art" than a science: for example, some ``best practices" include explicitly giving a role or persona to the LLM (e.g., ``you are an expert in \emph{X}"), with the idea that these instructions will condition the model to respond according to those instructions. Another example is ``chain of thought reasoning" \citep{wei2022chain}, or asking the model to think ``step-by-step", which has been surprisingly effective, and some have even likened this to human ``System 1/System 2" (intuitive vs. deliberative) thinking \citep{hagendorff2023human, yao2023tree}. But this also means that many prompts may, for unknown reasons, produce lower-quality output, which may lead to false negative inferences: one might argue that some paper's failure to get GPT-4 to perform \emph{X}, is because they did not find the ``right" prompt. 

On the one hand, humans also respond differently based on how questions are phrased, and that is no surprise to cognitive scientists, especially those with a social psychology background. On the other hand, this clashes with our mental model of how LLMs (and ``computer programs" more generally) work. 
%
%
%
%
%
The sensitivity (or less charitably, brittleness) of LLM performance to prompts suggests that as a first step, we need to \textbf{report prompts and procedures in full} (including hyperparameters: temperature; date on which the model was accessed for proprietary models, etc.). We might even need to do additional steps like \textbf{permuting answer choices on multiple-choice surveys}, or \textbf{permuting the order of information in presented stimuli}. As a field, we have lots of experience with such controlled experiments with humans (e.g., counterbalancing a blocked presentation design or counterbalancing the order of presentation of stimuli). There needs to be a similar paradigm shift---and field-wide discussions---when dealing with LLMs. 
We need to continually revise our scientific  conventions.

But at a broader level, the brittleness of LLMs to seemingly irrelevant changes in prompts, especially those that would not meaningfully affect humans, is concerning. Mathematically, it suggests that the model is overfitting. Cognitively, it suggests a learnt stimulus-response (memorization), rather than true conceptual understanding. Should we be inferring such complex cognition when we still lack an understanding of these boundary conditions (which differ from humans)?

\subsection{6. Generalizability and longevity of Results}

In science we often want to produce knowledge that is generalizable and ``true"---at least until future experiments falsify our theories. But, as the current review shows, many papers are making inferences based on the capabilities of currently-released LLM models. These models, in all likelihood and given the recent history of face-paced development, will be updated and perhaps made obsolete in a matter of months. Would then the corresponding results and inferences made on these models, also be made obsolete? (At a practical level, this also matters given the long review time in publishing). It is perhaps worth thinking about whether we as a field should consider prioritizing research paradigms and agendas that produce more generalizable, lasting knowledge that will last more than a few months.

\section{Conclusion}


It has been an exciting 2023--2024 for cognitive science, with many papers and preprints jumping on the opportunity to study these amazingly capable models, and the scientific inferences that we can glean from them. Indeed, these models are pushing the boundaries of our understanding of cognition, allowing more creative experiments with larger data, and increasing the external validity of our science.

However, as this review points out, there are still many challenges and issues that underlie these scientific endeavors. We have tried to briefly outline what types of inferences can be licensed with such evidence, and what are concerns that might undermine such inferences. 
We also note that most of the work has been on making inferences about LLM abilities---these inferences might be transient anyway, as models keep improving.
We hope that with more time and as these research paradigms mature, we can draw more insights about human cognition. This paper is not meant to provide a definitive framing of the field, but rather to start conversations about the outstanding issues in these new research endeavors, and we hope that it will succeed in doing so.



\section{Acknowledgments}

We would like to thank Robert Hawkins and Judith Fan for conversations that led to this paper, and four anonymous reviewers for their helpful feedback.


\printbibliography 

\end{document}